\documentclass[runningheads]{llncs}

\usepackage{eccv}

\usepackage{eccvabbrv}

\usepackage{graphicx}
\usepackage{booktabs}

\usepackage[accsupp]{axessibility}  

\usepackage{hyperref}
\hypersetup{
    colorlinks=true,
    citecolor=green,
    linkcolor=red,
    urlcolor=cyan,
}

\usepackage{orcidlink}

\usepackage{array}
\makeatletter
\newcommand{\printfnsymbol}[1]{%
  \textsuperscript{\@fnsymbol{#1}}%
}
\makeatother

\begin{document}


\title{TCLC-GS: Tightly Coupled LiDAR-Camera Gaussian Splatting for Autonomous Driving} 

\titlerunning{TCLC-GS}

\author{
Cheng Zhao\inst{1}\thanks{Equally contributed as co-first author.} \and
Su Sun\inst{2}\printfnsymbol{1} \and
Ruoyu Wang\inst{1} \and
Yuliang Guo\inst{1} \and
Jun-Jun Wan\inst{3} \and \\
Zhou Huang\inst{3} \and
Xinyu Huang\inst{1} \and
Yingjie Victor Chen\inst{2} \and
Liu Ren\inst{1}
}

\authorrunning{C. Zhao et al.}


\institute{Bosch Research North America \& Bosch Center for Artificial Intelligence (BCAI) 
\email{\{cheng.zhao, ruoyu.wang, yuliang.guo, xingyu.huang, liu.ren\}@us.bosch.com}
\and
Purdue University  ~~~~~\inst{3} XC Cross Domain Computing, Bosch \\
\email{\{sun931, victorchen\}@purdue.edu}  \email{\{jun-jun.wan, zhou.huang\}@bosch.com}
\\
}

\maketitle
\begin{abstract}
Most 3D Gaussian Splatting (3D-GS) based methods for urban scenes initialize 3D Gaussians directly with 3D LiDAR points, which not only underutilizes LiDAR data capabilities but also overlooks the potential advantages of fusing LiDAR with camera data.
In this paper, we design a novel tightly coupled LiDAR-Camera Gaussian Splatting~(TCLC-GS) to fully leverage the combined strengths of both LiDAR and camera sensors, enabling rapid, high-quality 3D reconstruction and novel view RGB/depth synthesis.
TCLC-GS designs a hybrid explicit (colorized 3D mesh) and implicit (hierarchical octree feature) 3D representation derived from LiDAR-camera data, to enrich the properties of 3D Gaussians for splatting.
3D Gaussian's properties are not only initialized in alignment with the 3D mesh which provides more completed 3D shape and color information, but are also endowed with broader contextual information through retrieved octree implicit features.
During the Gaussian Splatting optimization process, the 3D mesh offers dense depth information as supervision, which enhances the training process by learning of a robust geometry.
Comprehensive evaluations conducted on the Waymo Open Dataset and nuScenes Dataset validate our method's state-of-the-art~(SOTA) performance.
Utilizing a single NVIDIA RTX 3090 Ti, our method demonstrates fast training and achieves real-time RGB and depth rendering at 90 FPS in resolution of 1920$\times$1280 (Waymo),  
and 120 FPS in resolution of 1600$\times$900 (nuScenes) in urban scenarios. 
\end{abstract}

\section{Introduction}
\label{sec:introduction}
\begin{figure*}[!t]
    \centering
    \includegraphics[width=1.0\textwidth]{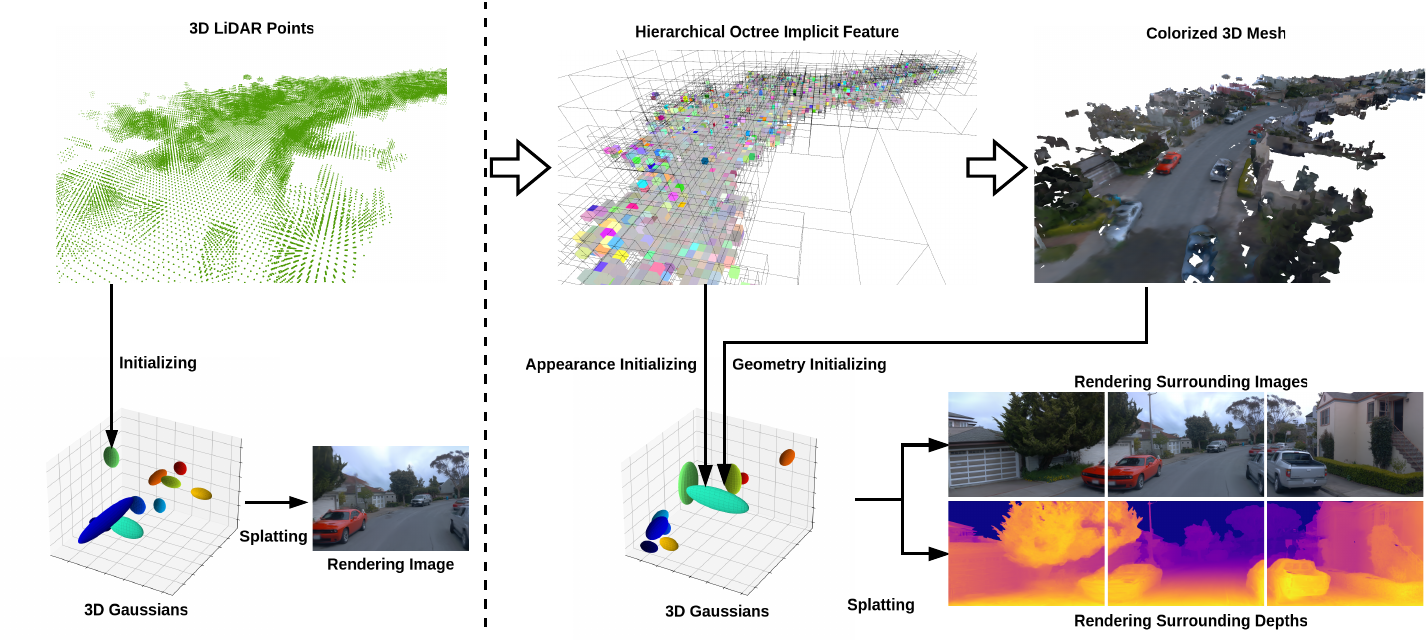}
    \caption{\textbf{Left: Original 3D-GS~\cite{kerbl20233d}} based methods directly initialize 3D Gaussians by 3D LiDAR points;
    \textbf{Right: Our TCLC-GS} enriches the geometry and appearance attributes of 3D Gaussians by explicit (colorized 3D mesh) and implicit (hierarchical octree feature) representations. }
  \label{fig:overview}
\end{figure*}
Urban-level reconstruction and rendering present significant challenges due to the vast scale of the unbounded environments and the sparse nature of the captured data. 
Fortunately, in autonomous vehicle settings, data from various modalities captured by multiple sensors are typically available. 
However, fully leveraging different modality data from multi-sensors for precise modeling and real-time rendering in urban scenes remains an open question in the field.

Neural Radiance Fields (NeRF)~\cite{mildenhall2021nerf}  based solutions are effective in reconstructing urban environments when a sufficient number of images captured from diverse viewpoints are available.
NeRF-W~\cite{martin2021nerf} and MipNeRF-360~\cite{barron2022mip} focused on utilizing NeRF for reconstructing street scenes within large-scale, unbounded scenarios. 
A further extension, Block-NeRF~\cite{tancik2022block}, adapts NeRF for city-level scenarios, constructing block-wise radiance fields from individual NeRF models to form a complete scene. 
However, the quality of modeling and rendering significantly deteriorates when relying solely on images captured from very sparse viewpoints.
In addition, a significant limitation of these methods is their dependence on intensive volumetric sampling in free space, leading to excessive consumption of computational resources in areas with no content.

The point-based urban radiance field methods~\cite{rematas2022urban, ost2022neural} integrate LiDAR point clouds to provide supervision, thereby facilitating geometry learning. 
These approaches take advantage of explicit 3D geometry derived from LiDAR data to represent the radiance field, enhancing rendering efficiency. 
Mesh-based rendering technique~\cite{lu2023urban, liu2023real} employs learned neural descriptors on the 3D mesh to achieve both accurate reconstruction and fast rasterization. 
Fusing these two modalities offers a viable solution to address the viewpoint sparsity in large-scale and complex scenes.
However, these methods face significant limitations due to their slow training and rendering speeds, which are further compounded by the critical requirement in real-time applications.

In contrast to Neural Radiance Fields~(NeRF), 3D Gaussian Splatting (3D-GS)~\cite{kerbl20233d}, utilizes a more extensive 3D Gaussian representation for scenes, which not only achieves faster training but also facilitates real-time rendering. 
The initial 3D-GS method sought to initialize Gaussians using the points from structure-from-motion (SfM).
However, this approach faces challenges in unbounded urban scenes within autonomous driving contexts, particularly when viewpoints are sparse. 
This sparsity can cause SfM techniques to fail in accurately and completely recovering scene geometries.
To facilitate better 3D Gaussian initialization, pioneering research~\cite{zhou2023drivinggaussian, yan2024street, chen2023periodic} has introduced LiDAR priors into the 3D-GS process, enabling more accurate geometries and ensuring rendering consistency of multiple surrounding views.
However, directly initializing 3D Gaussians' positions with LiDAR points
does not thoroughly exploit the rich 3D geometric information embedded within 3D points, such as depth and geometric features.

In this paper, we proposed a novel Tightly Coupled LiDAR-Camera Gaussian Splatting~(TCLC-GS) for accurate modeling and real-time rendering in surrounding autonomous driving scenes. 
Contrasting with the intuitive method of directly initializing 3D Gaussians using LiDAR points~(Fig.~\ref{fig:overview}.~Left), TCLC-GS~(Fig.~\ref{fig:overview}.~Right),  offers a more cohesive solution, effectively leveraging the combined strengths of both LiDAR and camera sensors.
The TCLC-GS's key idea is a hybrid 3D representation combining explicit (colorized 3D mesh) and implicit (hierarchical octree feature) derived from LiDAR-camera data, to enhance both geometry and appearance properties of 3D Gaussians. 
To be specific, we first learn and store implicit features in an octree-based hierarchical structure through encoding LiDAR geometries and image colors.
Then we initialize 3D Gaussians in alignment with a colorized 3D mesh decoded from the implicit feature volume. 
The 3D mesh enhances continuity/completeness, increases density, and adds color details compared to the original LiDAR points.
Meanwhile, we enhance the learning of appearance descriptions for each 3D Gaussian by incorporating implicit features retrieved from the octree.
We further render dense depths from the explicit meshes to supervise the GS optimization process, enhancing the training robustness compared to using sparse LiDAR depths.
By this way, LiDAR and camera data are tightly integrated within the initialization and optimization phases of 3D Gaussians.

The novel features of TCLC-GS can be summarized as follows:
1) Hybrid 3D representation provides a explicit~(colorized 3D mesh) and implicit~(hierarchical octree feature) representation to guide the properties initialization and optimization of 3D Gaussians;
2) The geometry attribute of 3D Gaussian is initialized to align with the 3D mesh which offers completed 3D shape and color information, and the appearance attribute of 3D Gaussian is enriched with retrieved octree implicit features which provides more extensive context information;
3) Besides RGB supervision, the dense depths rendered from the 3D mesh offer supplementary supervision in GS optimizations.
Our solution improves the quality of 3D reconstruction and rendering in urban driving scenarios, without compromising the efficiency of 3D-GS.
It enables us to fast and accurately reconstruct an urban street scene, while also achieving real-time RGB and depth rendering capabilities at around 90 FPS for a resolution of 1920$\times$1280, and around 120 FPS for a resolution of 1600 $\times$ 900 using a single NVIDIA GeForce RTX 3090 Ti.
\section{Related Work}
\label{sec:related_work}
NeRF-based techniques~\cite{martin2021nerf, barron2022mip, tancik2022block} have demonstrated significant effectiveness in large-scale urban scenarios for autonomous driving. 
NeRF-W~\cite{martin2021nerf} integrates frame-specific codes within its rendering pipeline, effectively managing photometric variations and transient objects.
Mip-NeRF 360~\cite{barron2022mip}, an advancement of Mip-NeRF~\cite{barron2021mip}, adapts to unbounded scenes by compressing the entire space into a bounded area, thereby enhancing the representativeness of position encoding. 
Block-NeRF~\cite{tancik2022block} tackles the challenge of modeling large-scale outdoor scenes by partitioning the target urban scene into several blocks, with each segment represented by an individual NeRF network,  boosting the overall modeling capability.

Point-based rendering techniques~\cite{rematas2022urban, ost2022neural, lu2023urban, liu2023real} are characterized by their use of learned neural descriptors on point clouds, coupled with differentiable rasterization via a neural renderer. 
Urban Radiance Field~\cite{rematas2022urban} employs LiDAR point clouds for supervision to facilitate geometry learning.
The Neural Point Light Field (NPLF)~\cite{ost2022neural} method leverages explicit 3D reconstructions derived from LiDAR data to efficiently represent the radiance field in rendering processes. 
DNMP~\cite{lu2023urban} employs learned neural descriptors on point clouds and marries this approach with differentiable rasterization, facilitated through a neural renderer.
NeuRas~\cite{liu2023real} uses a scaffold mesh as its input and optimizes a neural texture field to achieve rapid rasterization.

3D Gaussian Splatting (3D-GS)~\cite{kerbl20233d} establishes a set of anisotropic Gaussians within a 3D world and further performs adaptive density control to achieve real-time rendering results based on point cloud input.
Most recent research~\cite{zhou2023drivinggaussian, yan2024street, chen2023periodic} has expanded upon the original 3D-GS by integrating temporal/time cues to model the dynamic objects in the urban environments. 
However, these methods only take advantage of 3D LiDAR points to initialize the position of 3D Gaussians, which does not entirely fulfill the potential for an efficient LiDAR-camera fusion solution.
\section{Methodology}
\label{sec:methodology}
\begin{figure*}[!t]
    \centering
    \includegraphics[width=1.0\textwidth]{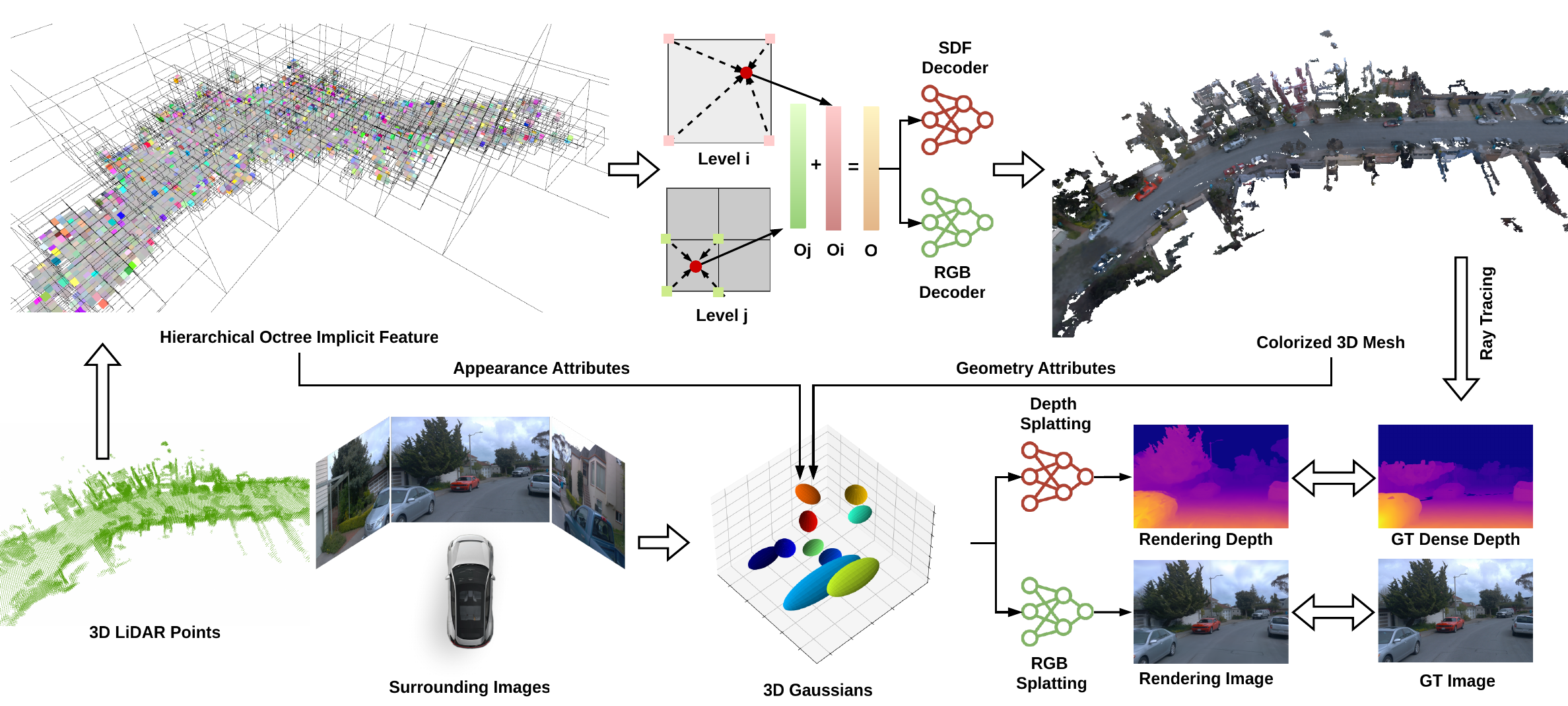}
    \caption{\textbf{The pipeline of TCLC-GS:}
    We first merge all the LiDAR sweeps together, and then build a hierarchical octree implicit feature grid using the sampled 3D point within the truncation region along the LiDAR rays.
    These octree implicit features are trained with SDF and RGB decoders, supervised by sparse LiDAR range measurements and surrounding image projected RGB colors. 
    Subsequently, we obtain the optimized octree implicit representations and colorized 3D mesh of the global scene.
    The geometry attributes of 3D Gaussians are initialized by the 3D mesh
    while the appearance attributes of 3D Gaussians are enriched by the mesh-vertex-retrieved octree implicit features. 
    The 3D Gaussians are optimized through depth and RGB splatting with dense depth and color supervision. 
    Different from the sparse depth supervision derived from LiDAR, our dense depth supervision is rendered from the 3D mesh utilizing the Ray Tracing method.
    }
  \label{fig:pipeline}
\end{figure*}

\textbf{Overview:} 
The pipeline for our TCLC-GS is detailed in Fig.~\ref{fig:pipeline}. 
The TCLC-GS framework is composed of two primary learning components:  
1) the octree implicit feature with SDF and RGB decoders (Fig.~\ref{fig:pipeline} upper part), 
and 2) 3D Gaussians with depth and RGB splattings (Fig.~\ref{fig:pipeline} lower part).
The LiDAR and camera data are tightly integrated in a uniform framework.

\textbf{Problem Definition:}
In an urban street scene, given a sequence of surrounding images and LiDAR data collected from a vehicle-mounted system, our goal is to develop a model to reconstruct the environment and generate photorealistic images and depths from novel viewpoints. 
A set of surrounding images $\mathcal{I}$ are captured by multiple surrounding cameras with corresponding intrinsic matrices $I$ and extrinsic matrices $E$.
A set of 3D points $\mathcal{P}$ captured by LiDAR with corresponding extrinsic matrices $E'$.
Meantime, the vehicle trajectory $\mathcal{T}$ is provided or calculated by multi-sensor-based odometry. 
Our model $\mathcal{R}$ aims to achieve precise 3D reconstruction and synthesize novel
viewpoints at novel camera pose $[E_n, I_n]$ by rendering $\hat{\mathcal{I}}, \hat{\mathcal{D}} = \mathcal{R}(E_n, I_n)$.

\textbf{Hierarchical Octree Implicit Feature:}
To encapsulate fine-grained geometric details and contextual structure information of the global 3D scene, we learn and store implicit features through learnable octree-based hierarchical feature grids $\mathcal{F}$ similar to NGLOD~\cite{takikawa2021nglod} from a sequence of LiDAR data.
The implicit features can be further decoded into signed distance values (SDFs) and RGB colors through a shallow dual-branch MLP decoder. 

We first merge a sequence of LiDAR sweeps using vehicle trajectory $\mathcal{T}$ to obtain the global point cloud of the scene.
We construct a $L$ level octree using the point clouds, and store latent features at eight corners per octree node in each tree level.
Multi-resolution features are stored across different octree levels, where the octree feature at level $i$ is denoted as $F_{i} \in \mathcal{F}$. 
The fine-grained features at the lower levels of the octree capture detailed geometry, whereas the coarse features at the higher levels encode broader contextual information.
To optimize the implicit features, we sample 3D query points $p\in \mathbb{R}^3$ within the truncation region of beam endpoints along the LiDAR ray. 
For a query point $p$, we compute its feature vector $F_{i}(p)$ by trilinear interpolating its corresponding features at octree node corners.
We feed both position encoding and multi-resolution octree features retrieved by a query point $p$ into the dual branch MLP decoder, i.e., SDF decoder $D_{s}$ to estimate the SDF $\hat{s}_{p}$, and RGB decoder $D_{c}$ to estimate the RGB color $\hat{c}_{p}$, 
\begin{equation}
\hat{s}_{p} = D_{s}(f(p), F_i(p)), \hat{c}_{p} = D_{c}(f(p), F_i(p)), i = 1,2,...L
\label{eq:1}
\end{equation}
where $f$ refers to the position encoding function.

We train $D_{s}$ using binary cross entropy (BCE) loss $\mathcal{L}_{bce}$ as,
\begin{equation}
\mathcal{L}_{bce}(p) = S(s_p) \cdot log(S(\hat{s}_{p})) + (1-S(s_p)) \cdot log(1-S(\hat{s}_{p})),
\label{eq:2}
\end{equation}
where $S(x) = 1/(1+e^{x/\sigma})$ is sigmoid function with a flatness hyperparameter $\sigma$.
The ground truth $s_p$ is calculated by the signed distance between the sampled point to the beam endpoint along the LiDAR ray.
We further employ two regularization terms together with the BCE loss: eikonal term $\mathcal{L}_{eik}$~\cite{icml2020_2086} and smoothness term $\mathcal{L}_{smooth}$~\cite{Ortiz:etal:iSDF2022},
\begin{equation}
\label{eq:3}
\mathcal{L}_{eik}(p) = (1 - || \nabla D_{s}(f(p), F_i(p)) ||)^2 ,
\end{equation}
\begin{equation}
\label{eq:4}
\mathcal{L}_{smooth}(p) = || \nabla D_{s}(f(p), F_i(p)) - \nabla D_{s}(f(p+\epsilon), F_i(p+\epsilon)) ||^2,
\end{equation}
where $\epsilon$ is a small perturbation.

We train $D_{c}$ using $\mathcal{L}_1(p) = | \hat{c}_{p} - c_{p} |$ loss. 
For each LiDAR point $p$, we obtain its RGB ground truth $c_p$ by projecting the point to the camera image plane using calibration matrices $E', E, I$.
A LiDAR point might be projected onto multiple pixels across multiple surrounding images. 
We select the RGB of pixel from the image plane with the smallest Euclidean distance to the point as ground truth.

The global objective function $\mathcal{L}(p)$ of octree implicit feature learning is defined as,
\begin{equation}
\label{eq:6}
\mathcal{L}_(p) = \mathcal{L}_{bce}(p) + \lambda_{eik} \mathcal{L}_{eik}(p) + \lambda_{smooth} \mathcal{L}_{smooth}(p) + \lambda_{RGB} \mathcal{L}_{1}(p),
\end{equation}
where $\lambda_{eik}$, $\lambda_{smooth}$ and $\lambda_{RGB}$ are scale factors. We randomly initialize the corner features when creating the octree and optimizing them during training. 
Through encoding both LiDAR geometries and image color into an octree representation of the entire scene, 
we enrich the of 3D Gaussian's properties with high-dimensional implicit features, as detailed in the subsequent steps. 
\begin{figure*}[!t]
    \centering
    \includegraphics[width=1.0\textwidth]{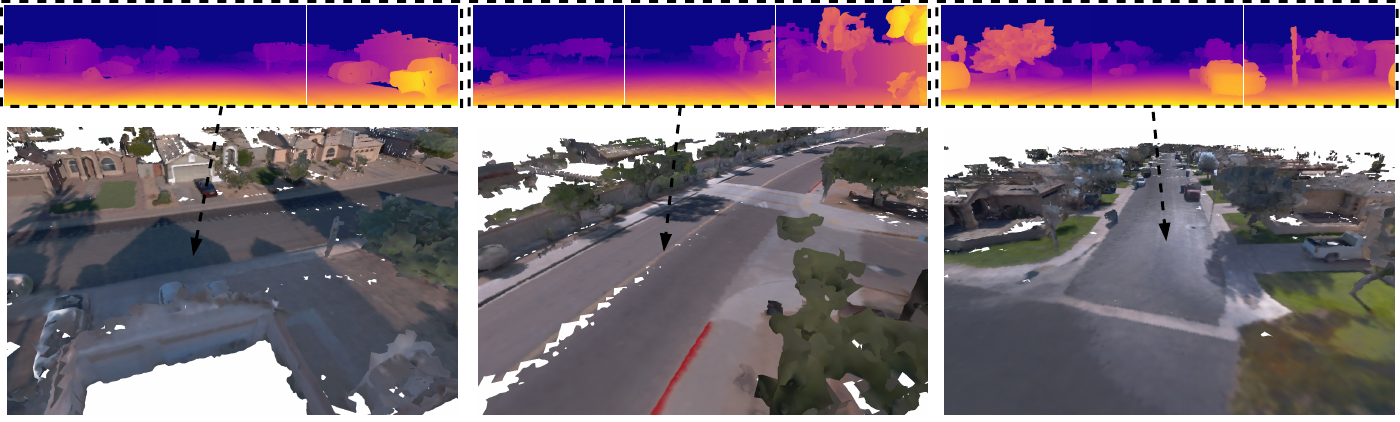}
    \caption{\textbf{Visualization of our colorized 3D mesh and dense depths.}
    Row 1: rendered dense surrounding depth images given the camera pose within the 3D mesh;
    Row 2: generated colorized 3D mesh based on the octree implicit representation. 
    }
  \label{fig:mesh_depth}
\end{figure*}

\textbf{3D Mesh and Dense Depth Generation:}
After learning and storing the octree implicit features, which includes both geometry and RGB information, we can generate a colorized 3D mesh.
Specifically, we first uniformly sample 3D points within the scene's 3D spatial space.
Each of these points is then used to retrieve octree implicit features, which are then fed into a dual-branch decoder.
Subsequently, we generate a colorized 3D mesh $\mathcal{M}$ in the form of a triangle mesh by marching cubes~\cite{lorensen1987marching} based on the decoder predicted SDFs and colors. 
Furthermore, from this 3D mesh $\mathcal{M}$, we generate a set of dense depth images $\mathcal{D}$ along the vehicle trajectory by Ray Tracing~\cite{glassner1989introduction} as, 
\begin{equation}
\label{eq:7}
\mathcal{D} = RayTracing(\mathcal{M} | \mathcal{T}, E, I, E').
\end{equation}
We present visualization examples of the colorized 3D mesh and dense depths generated by our method in Fig.~\ref{fig:mesh_depth}.

Through representing the scene as a mesh, the learning of 3D Gaussians is enhanced with a more robust geometry and appearance prior.
In detail, the mesh fully exploits the geometries from LiDAR points in terms of continuity/completion and density. 
Compared with sparse raw LiDAR points, the mesh representations are able to reconstruct missed geometries due to sparse scans and viewpoints, and provides scalable and adaptive geometry priors.
Meanwhile, the colorized 3D mesh provides RGB information for any 3D point positions sampled on its surface. 
This is a significant improvement over the intuitive method of projecting LiDAR points onto surrounding images, which typically assigns colors to only about one-third of the LiDAR points due to limited sensor overlap between camera and LiDAR.
In contrast to the sparse depth obtained from LiDAR points, the dense depth rendered from the 3D mesh provides more robust supervision and alleviate the over-fitting during training.

\textbf{LiDAR-Camera Gaussian Splatting:}
3D-GS~\cite{kerbl20233d} employs a collection of 3D Gaussians to represent a scene. 
It uses a tile-based rasterization process, enabling real-time alpha blending of multiple Gaussians. 
Different from directly initializing Gaussian by LiDAR points, our method initials 3D Gaussians $\mathcal{GS}$ on the 3D mesh $\mathcal{M}$.
We follow the rule of Gaussians bound to the triangles~\cite{guedon2023sugar} to keep the 3D Gaussians flat and aligned with the mesh triangles.
Each vertex $v$ in the mesh provides position $\mu$, color $c$, and the vertex-retrieved octree implicit feature $F(v)$ to initialize the 3D Gaussians’ attributes as,
\begin{equation}
\mathcal{GS} = \{  \mu(v), s(v), q(v), c(v) | \mathcal{M}; o(F_i(v)), sh(c(v), F_i(v)), F_i(v) | \mathcal{M}, \mathcal{F} \}, i=1,..L.
\label{eq:8}
\end{equation}
The 3D Gaussians’ attributes include mean position $\mu \in \mathbb{R}^3 $, anisotropic covariance  $\Sigma \in \mathbb{R}^{3\times3}$ represented by a scale $s \in \mathbb{R}^3$ (diagonal matrix) and rotation $q$ (unit quaternion), opacity $o$ and spherical harmonics coefficients $sh$.
The vertices and colors of 3D mesh $\mathcal{M}$ serve as geometry and appearance priors for $\mu, s, q, c$ and $o, sh$ properties of 3D Gaussians. 
Additionally, the octree implicit features are also incorporated to improve appearance descriptions $o, sh$ of 3D Gaussians. 
Please note that the newly added properties $F_i(v)$ are dynamically controlled through cloning, and pruning alongside other attributes by adaptive density control during optimization. 
Additionally, to enhance training robustness, we use very shallow MLPs to encode the $c$ and $F_i(v)$ before their integration into the 3D Gaussians.

We utilize a differentiable 3D Gaussian splatting renderer to project each 3D Gaussian onto the 2D image plane, resulting in a collection of 2D Gaussians.
The covariance matrix $\Sigma'$ in camera coordinates is computed by, 
\begin{equation}
\Sigma' = JE \Sigma E^TJ^T,
\label{eq:9}
\end{equation}
where $E$ refers to the world-to-camera matrix and $J$ refers to Jacobian of the perspective projection $I$.

By sorting the Gaussians according to their depth within the camera space, we can effectively query the attributes of each 2D Gaussian.
This step facilitates the following volume rendering process to estimate the color $C$, depth $D$ and accumulated opacity $O$ of each pixel as,
\begin{equation}
C = \sum_{i=1}^{N} T_i \alpha_i c_i, D = \sum_{i=1}^{N} T_i \alpha_i z_i, O = \sum_{i=1}^{N} T_i \alpha_i, T_i = \prod_{j=1}^{i-1}(1 - \alpha_j),
\label{eq:10}
\end{equation}
where $z$ denotes the distance from the image plane to the Gaussian point center.
$\alpha$ is calculated by $\alpha_i = o_i \cdot exp(-\frac{1}{2} (x-\mu_i)^T \Sigma^{'-1} (x-\mu_i) )$

Given a specific view ($E, I$) in a scene containing $N$ 3D Gaussians, both the image $\hat{\mathcal{I}}$ and depth $\hat{\mathcal{D}}$ are rendered by the differentiable rendering function $\mathcal{R}$ as,
\begin{equation}
\hat{\mathcal{I}}, \hat{\mathcal{D}} = \mathcal{R}( \{\mathcal{GS}_i , i=1,2,..N\} | E, I ).
\label{eq:11}
\end{equation}

Note all depths used in this paper are inverse depths. 
The parameters of $\mathcal{GS}$ will be optimized during training by the objective function $\mathcal{L}$ as following,
\begin{equation}
\begin{split}
\mathcal{L} = (1-\lambda_{ssim}) \mathcal{L}_{1}(\mathcal{I}, \hat{\mathcal{I}}) + \lambda_{ssim} \mathcal{L}_{ssim}(\mathcal{I}, \hat{\mathcal{I}}) + \lambda_{depth} \mathcal{L}_{1}(\mathcal{D}, \hat{\mathcal{D}}) + \\ \lambda_{smooth} \mathcal{L}_{smooth}(\mathcal{I}, \hat{\mathcal{D}}) + \lambda_{sky} \mathcal{L}_{sky},
\end{split}
\label{eq:12}
\end{equation}
where $\lambda_{ssim}$, $\lambda_{depth}$, $\lambda_{smooth}$ and $\lambda_{sky}$ are scale factors. 
$\mathcal{I}$ refers to the original RGB image, and $\mathcal{D}$ refers to the dense depth rendered from 3D mesh by Ray Tracing.
$\mathcal{L}_{ssim}$ refers to Structural Similarity Index Measure~(SSIM) term between the original RGB and rendered RGB images, and $\mathcal{L}_{smooth}$ refers to inverse depth smooth~\cite{monodepth17} term between the original RGB image and rendered depth.
$\mathcal{L}_{sky}$ refers to the sky opacity loss~\cite{yang2023emernerf} defined as BCE loss,
\begin{equation}
\mathcal{L}_{sky} = \mathcal{L}_{bce}(O, 1-M_{sky})
\label{eq:13}
\end{equation}
where $M_{sky}$ is the sky mask from a pretrained segmentation model SegFormer~\cite{xie2021segformer}.
The $\mathcal{L}_{sky}$ drives the opacity of sky pixels toward zero to be transparent, which pushes the 3D Gaussian positions of the sky far away. 
The $\mathcal{L}_{sky}$ significantly reduces the sky artifacts, enabling the depth synthesis with clear sky areas.
These sky masks are only necessary during training and not during inference.

The enhancement of Gaussian optimization is achieved by improving the initialization of 3D Gaussians' geometry and appearance, enriching the appearance descriptions of 3D Gaussians, and integrating dense depth supervision.
In order to mitigate the scale ambiguity due to the long duration of the driving scenario, 
we employ the incremental optimization strategy~\cite{zhou2023drivinggaussian}, 
which involves incrementally adding 3D Gaussians by uniformly segmenting the scenario into a series of bins based on the LiDAR depth range.
For each bin, we further employ position-aware point adaptive control~\cite{chen2023periodic} during Gaussian optimization, which utilizes smaller points for nearby positions and larger points for distant locations in the unbounded scene. 
\section{Experiments}
\label{sec:experiments}
\begin{figure*}[!t]
    \centering
    \includegraphics[width=1.0\textwidth]{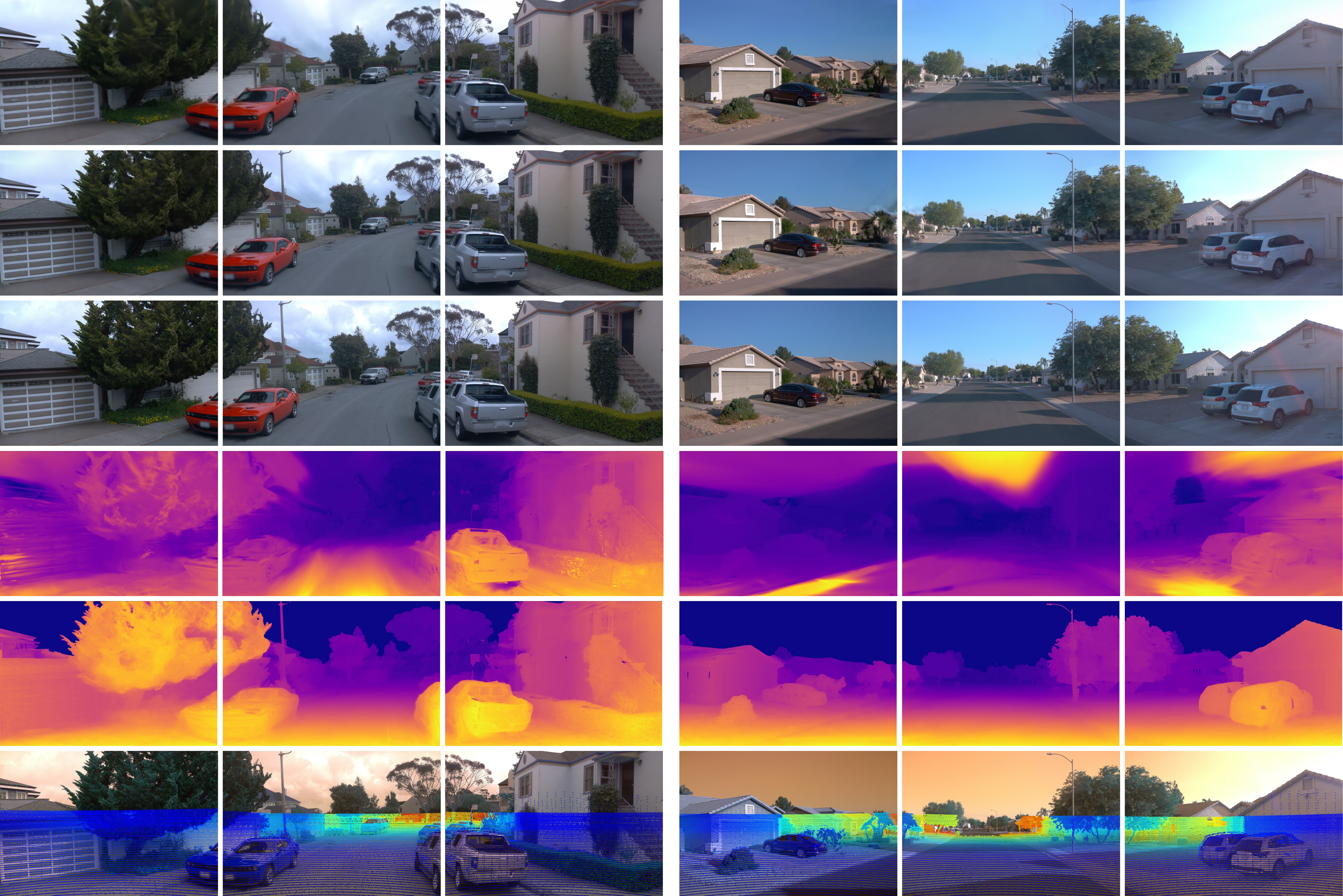}
    \caption{Visual comparison of image and depth synthesis from novel front-left, front, and front-right surrounding views on the Waymo dataset. 
    Row 1: 3D-GS images;
    Row 2: TCLC-GS images;
    Row 3: GT images;
    Row 4: 3D-GS depths;
    Row 5: TCLC-GS depths;
    Row 6: GT depth of LiDAR points projected on images.}
  \label{fig:waymo_vis}
\end{figure*}
\textbf{Datasets:}
Our experimental evaluations are conducted on two of the most widely-used datasets in autonomous driving research: the Waymo Open Dataset \cite{sun2020scalability} and the nuScenes Dataset~\cite{caesar2020nuscenes}. 
The Waymo Open Dataset offers urban driving scenarios, with each scenario lasting 20 seconds and recorded using five high-resolution LiDARs and five cameras oriented towards the front and sides. 
For our experiments, we selected six challenging recording sequences from this dataset, utilizing surrounding views captured by three cameras and corresponding data from five LiDAR sweeps. 
Each sequence consists of approximately 100 frames, and we use a random one of every tenth frame as a test frame and the remaining for training.
The nuScenes dataset, a large-scale public resource for autonomous driving research, includes data from an array of sensors: six cameras, one LiDAR, five RADARs, GPS, and IMU. 
In our experiments, we use the keyframes from six challenging scenes, which include surrounding views from six cameras and corresponding LiDAR sweeps.
In each sequence, which comprises roughly 40 frames, we randomly select one of every fifth frame as a test frame and utilize the rest for training purposes.
To comprehensively evaluate and compare the detail synthesis capabilities, we train and assess our methods and all baselines using full-resolution images,
i.e., 1920$\times$1280 for the Waymo dataset and 1600$\times$900 for the nuScenes dataset.

\newcolumntype{C}[1]{>{\centering\arraybackslash}p{#1}}
\begin{table*}[t]
\centering
\resizebox{0.999\textwidth}{!}{
\begin{tabular}{C{3cm} C{5cm} C{5cm}}
\hline
\textbf{Sequence} & \begin{tabular}[c]{@{}c@{}}3D-GS~\cite{kerbl20233d}\\ PSNR↑/SSIM↑/LPIPS↓\end{tabular} & \begin{tabular}[c]{@{}c@{}}TCLC-GS\\ PSNR↑/SSIM↑/LPIPS↓\end{tabular} \\ 
\hline
Segment-1024795...    & 26.19/0.84/0.25 & 28.28/0.89/0.16\\ 
Segment-1071392...    & 27.35/0.84/0.25 & 28.68/0.88/0.17 \\ 
Segment-1103765...    &26.48/0.67/0.44 & 28.54/0.71/0.42 \\
Segment-1346990...    & 25.08/0.84/0.26 & 27.72/0.88/0.16\\
Segment-1433374...    & 26.20/0.85/0.25 & 27.30/0.87/0.21 \\
Segment-1466335...    & 26.87/0.86/0.25 & 28.13/0.90/0.18 \\
\hline
Average               & 26.36/0.82/0.28 & \textbf{28.11/0.86/0.22} \\
\hline
\end{tabular}}
\caption{Performance comparison of image synthesis from novel views between the proposed method and baseline on the Waymo dataset.}
\label{tab:table1}
\end{table*}
%
\begin{table*}[t]
\centering
\resizebox{0.999\textwidth}{!}{
\begin{tabular}{C{3cm} C{5cm} C{5cm}}
\hline
\textbf{Sequence} & \begin{tabular}[c]{@{}c@{}}3D-GS~\cite{kerbl20233d}\\ AbsRel↓/RMSE↓/RMSElog↓\end{tabular} & \begin{tabular}[c]{@{}c@{}}TCLC-GS\\ AbsRel↓/RMSE↓/RMSElog↓\end{tabular} \\ 
\hline
Segment-1024795...    & 0.37/6.66/1.10 & 0.03/0.77/0.05 \\ 
Segment-1071392...    & 0.22/7.80/0.34 & 0.03/3.10/0.08 \\ 
Segment-1103765...    & 0.55/16.37/1.36 & 0.02/1.18/0.04 \\
Segment-1346990...    & 0.34/7.31/0.77 & 0.02/1.01/0.04 \\
Segment-1433374...    & 0.48/12.75/1.37 & 0.03/1.56/0.05 \\
Segment-1466335...    & 0.56/14.62/1.43 & 0.03/1.60/0.06 \\
\hline
Average               & 0.42/10.92/1.06 & \textbf{0.03/1.54/0.05} \\
\hline
\end{tabular}}
\caption{Performance comparison of depth synthesis from novel views between the proposed method and baseline on the Waymo dataset.}
\label{tab:table2}
\end{table*}

\textbf{Metrics:}
Following the previous research~\cite{ost2022neural, lu2023urban, zhou2023drivinggaussian, yan2024street, chen2023periodic}, our image synthesis evaluation employs three widely-used benchmark metrics, i.e., peak signal-to-noise ratio (PSNR), structural similarity index measure (SSIM), and the learned perceptual image patch similarity (LPIPS) in novel views.
Similarly as the previous research~\cite{bhat2023zoedepth, yang2024depth}, we choose three widely-used benchmark metrics, i.e., Absolute Relative Difference (AbsRel), Root Mean Squared Error (RMSE) and Root Mean Squared Error in the Logarithmic Scale (RMSElog)
for depth synthesis evaluation from novel views.
The ground truth RGB images are provided directly from the datasets, while sparse depth images, used as depth ground truth for depth evaluation, are obtained by projecting LiDAR points onto the surrounding images according to the calibration information.

\begin{figure*}[!t]
    \centering
    \includegraphics[width=1.0\textwidth]{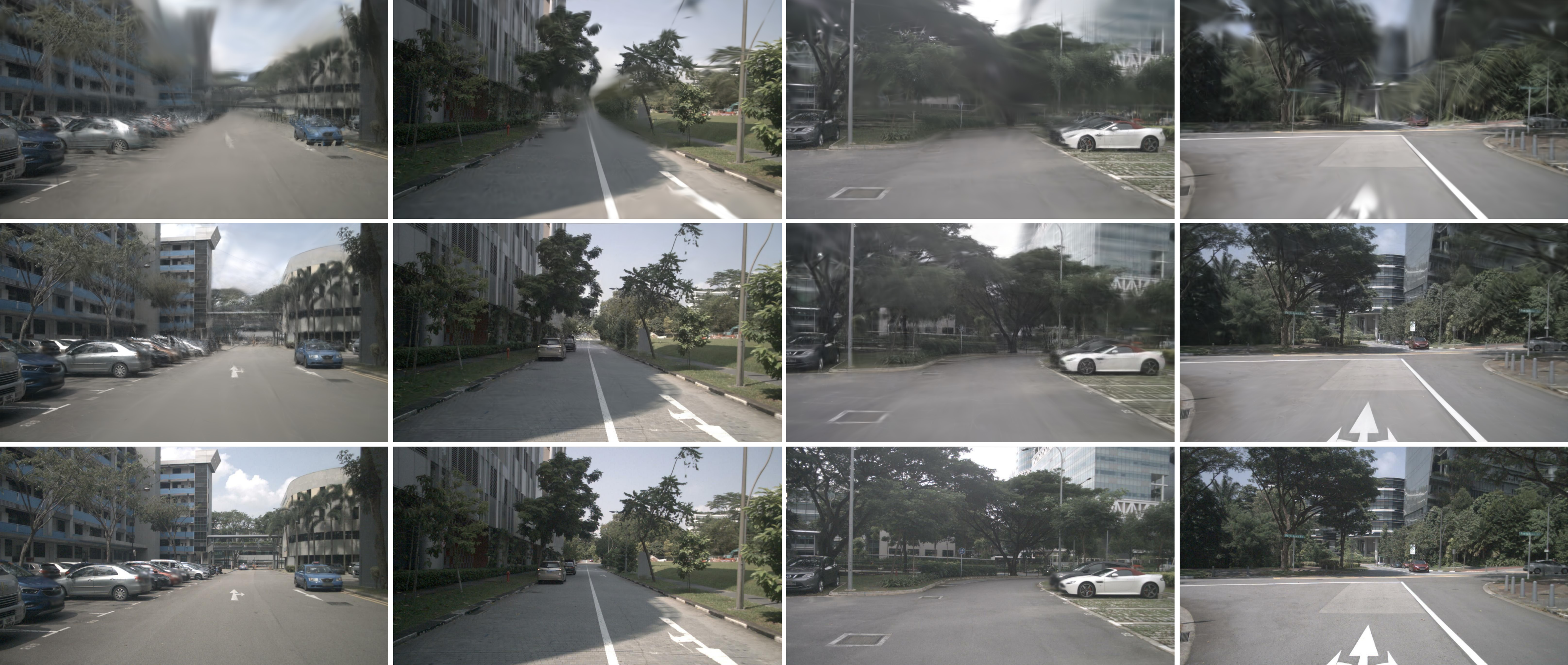}
    \caption{Visual comparison of image synthesis from novel views on nuScenes dataset. 
    Row 1: 3D-GS;
    Row 2: TCLC-GS;
    Row 3: GT.}
  \label{fig:nuScenes_image}
\end{figure*}
\begin{table*}[t]
\centering
\resizebox{0.999\textwidth}{!}{
\begin{tabular}{C{3cm} C{5cm} C{5cm} }
\hline
\textbf{Sequence} & \begin{tabular}[c]{@{}c@{}}3D-GS~\cite{kerbl20233d}\\ PSNR↑/SSIM↑/LPIPS↓\end{tabular} & \begin{tabular}[c]{@{}c@{}}TCLC-GS\\ PSNR↑/SSIM↑/LPIPS↓\end{tabular} \\ 
\hline
Scene-0008    & 25.32/0.82/0.26 & 26.41/0.85/0.22 \\
Scene-0051    & 25.84/0.85/0.25 & 26.78/0.86/0.23 \\ 
Scene-0058    & 24.71/0.81/0.27 & 26.45/0.88/0.20 \\
Scene-0062    & 25.91/0.87/0.22 & 27.47/0.89/0.17 \\
Scene-0129    & 26.87/0.88/0.21 & 28.65/0.90/0.16 \\ 
Scene-0382    & 23.96/0.82/0.32 & 25.76/0.84/0.26 \\ 
\hline
Average       & 25.44/0.84/0.26 & \textbf{26.92/0.87/0.21} \\
\hline
\end{tabular}}
\caption{Performance comparison of image synthesis from novel views between the proposed method and baseline on the nuScenes dataset.}
\label{tab:table4}
\end{table*}

\textbf{Baselines:}
We selected 3D-GS~\cite{kerbl20233d}, which utilizes LiDAR points to directly initialize 3D Gaussians, as our primary baseline for comparison. 
We also selected 
NeRF~\cite{mildenhall2021nerf},       
NeRF-W~\cite{martin2021nerf},      
Instant-NGP~\cite{muller2022instant},  
Point-NeRF~\cite{xu2022point},   
NPLF~\cite{ost2022neural},         
Mip-NeRF~\cite{barron2021mip},    
Mip-NeRF 360~\cite{barron2022mip},
DNMP~\cite{lu2023urban},        
as the supplementary baselines for further comparison.

\textbf{Evaluation on Waymo Open Dataset:}
We evaluated the proposed method by comparing it with the baseline on the Waymo Open dataset. 
The performance comparison of image and depth synthesis from novel views, relative to the main baseline, is detailed in Table~\ref{tab:table1} and Table~\ref{tab:table2} separately.
Additional comparisons of image synthesis from novel views against a broader range of baselines are depicted in Table~\ref{tab:table3}.
As indicated in Table~\ref{tab:table1}, our method outperforms 3D-GS across individual scenes in terms of PSNR, SSIM, and LPIPS metrics for novel image synthesis.
Furthermore, Table~\ref{tab:table3} shows that our method exceeds a broader array of baselines in average performance across these same metrics (PSNR, SSIM, and LPIPS) for novel image synthesis.
Similarly, Table~\ref{tab:table2} demonstrates that our approach significantly surpasses 3D-GS in the depth synthesis metrics of AbsRel, RSME, and RSMElog.
The significant improvement in depth synthesis performance can be attributed to the robust supervision provided by the rendered dense depths derived from the generated accurate 3D mesh.
For a visual perspective, the comparison results of image synthesis from novel surrounding views are illustrated in Fig.~\ref{fig:waymo_vis}.
We can see that TCLC-GS renders more clear and accurate RGB images than the 3D-GS, especially on roadside objects and in distant areas viewed from the front, front-left, and front-right perspectives.
Similarly, the visual comparison results for depth synthesis from novel surrounding views are showcased in Fig.~\ref{fig:waymo_vis}.
Here, TCLC-GS is observed to render denser and sharper depths compared to 3D-GS, especially in areas further away in the front, front-left, and front-right views. 

\begin{figure*}[!t]
    \centering
    \includegraphics[width=1.0\textwidth]{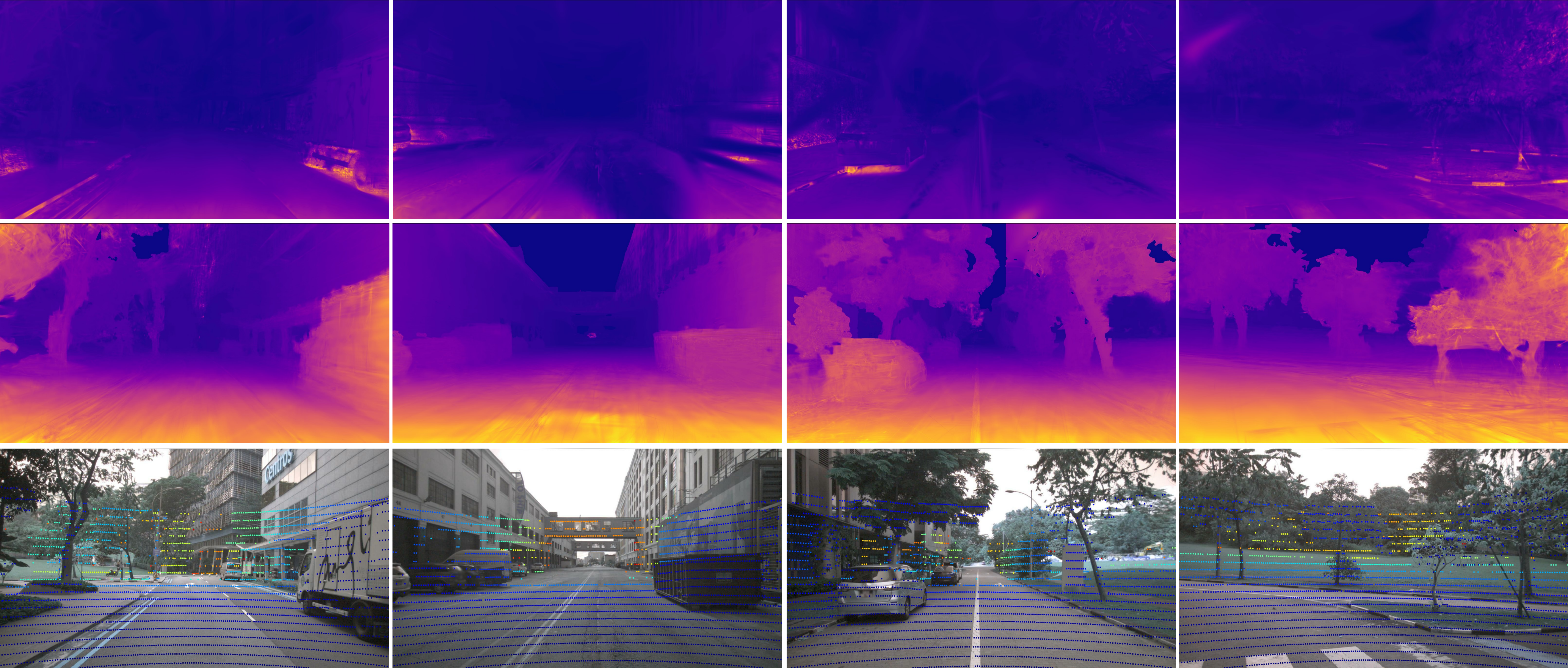}
    \caption{Visual comparison of depth synthesis from novel views on nuScenes dataset. 
    Row 1: 3D-GS;
    Row 2: TCLC-GS;
    Row 3: GT of LiDAR points projected on images.}
  \label{fig:nuScenes_depth}
\end{figure*}
\begin{table*}[t]
\centering
\resizebox{0.999\textwidth}{!}{
\begin{tabular}{C{3cm} C{5cm} C{5cm} }
\hline
\textbf{Sequence} & \begin{tabular}[c]{@{}c@{}}3D-GS~\cite{kerbl20233d}\\ AbsRel↓/RMSE↓/RMSElog↓\end{tabular} & \begin{tabular}[c]{@{}c@{}}TCLC-GS\\ AbsRel↓/RMSE↓/RMSElog↓\end{tabular} \\ 
\hline
Scene-0008    & 0.31/9.97/0.47 & 0.07/4.27/0.12 \\
Scene-0051    & 0.47/8.68/0.60 & 0.05/2.21/0.07 \\ 
Scene-0058    & 0.35/10.48/0.55 & 0.06/4.23/0.11 \\
Scene-0062    & 0.32/8.22/0.39 & 0.04/4.03/0.08 \\
Scene-0129    & 0.29/8.01/0.44 & 0.05/2.83/0.09 \\ 
Scene-0382    & 0.32/10.91/0.51 & 0.05/4.58/0.10 \\
\hline
Average       & 0.34/9.38/0.49 & \textbf{0.05/3.69/0.10} \\
\hline
\end{tabular}}
\caption{Performance comparison of depth synthesis from novel views between the proposed method and baseline on the nuScenes dataset.}
\label{tab:table5}
\end{table*}

\textbf{Evaluation on nuScenes Dataset:}
We conducted further comprehensive evaluations comparing our method with the baseline on the nuScenes dataset. 
The results of image and depth synthesis from novel viewpoints, compared with the primary baseline 3D-GS, are detailed in Table~\ref{tab:table4} and Table~\ref{tab:table5} separately. 
According to Table~\ref{tab:table4}, our method outperforms 3D-GS in the metrics of PSNR, SSIM, and LPIPS for novel image synthesis. 
Likewise, as indicated in Table~\ref{tab:table5}, our approach significantly excels beyond 3D-GS in the depth synthesis metrics of AbsRel, RSME, and RSMElog.
The lower depth performance of our method observed on the nuScenes dataset, compared to the Waymo dataset, is due to the nuScenes dataset's reliance on relatively sparse 32-line LiDAR data. 
In contrast, the Waymo dataset includes much denser LiDAR data, which contributes to a more accurate generation of 3D mesh and, subsequently, better rendered dense depth.
In Table~\ref{tab:table6}, we present a performance comparison between TCLC-GS and additional baselines on the nuScenes dataset, showcasing the average performance on PSNR, SSIM, and LPIPS metrics.
Our method surpasses these baselines in all mentioned metrics, demonstrating its superior performance in image synthesis.
Visual comparison results depicted in Figure~\ref{fig:nuScenes_image} and \ref{fig:nuScenes_depth} illustrate that our method not only produces clearer synthesized RGB images with fewer artifacts and less blurring but also generates sharper and denser synthesized depth images compared to the 3D-GS baseline.

\begin{table}[t]
    \centering
    \begin{minipage}{0.48\textwidth}
    \centering
    \resizebox{1.0\textwidth}{!}{
    \begin{tabular}{C{3cm} C{1cm} C{1cm} C{1cm}}
    \hline
    \textbf{Method}   & PSNR↑ & SSIM↑ & LPIPS↓     \\  
    \hline
    NeRF~\cite{mildenhall2021nerf}         & 26.24 & 0.87 & 0.47 \\
    NeRF-W~\cite{martin2021nerf}       & 26.92 & 0.89 & 0.42 \\
    Instant-NGP~\cite{muller2022instant}  & 26.77 & 0.88 & 0.40 \\
    Point-NeRF~\cite{xu2022point}   & 26.26 & 0.87 & 0.45 \\
    NPLF~\cite{ost2022neural}         & 25.62 & 0.88 & 0.45 \\
    Mip-NeRF~\cite{barron2021mip}     & 26.96 & 0.88 & 0.45 \\
    Mip-NeRF 360~\cite{barron2022mip} & 27.43 & \textbf{0.89} & 0.39 \\
    DNMP~\cite{lu2023urban}         & 27.62 & \textbf{0.89} & 0.38 \\
    3D-GS~\cite{kerbl20233d}     & 26.36 & 0.82 & 0.28 \\
    \hline
    TCLC-GS                      & \textbf{28.11} & 0.86 & \textbf{0.22}  \\ 
    \hline
    \end{tabular}}
    \caption{Performance comparison of image synthesis from novel view between the proposed method and additional comprehensive baselines on the Waymo dataset.}
    \label{tab:table3}
    \end{minipage}
    \hfill
    \begin{minipage}{0.48\textwidth}
    \resizebox{1.0\textwidth}{!}{
    \begin{tabular}{C{3cm} C{1cm} C{1cm} C{1cm}}
    \hline
    \textbf{Method}   & PSNR↑ & SSIM↑ & LPIPS↓     \\  
    \hline
    Mip-NeRF~\cite{barron2021mip}     & 18.10 & 0.63 & 0.46 \\
    Mip-NeRF 360~\cite{barron2022mip} & 23.45 & 0.74 & 0.35 \\
    S-NeRF~\cite{li2024scenarionet}   & 25.62 & 0.77 & 0.27 \\
    3D-GS~\cite{kerbl20233d}     & 25.44 & 0.84 & 0.26 \\
    \hline
    TCLC-GS                      & \textbf{26.92} & \textbf{0.87} & \textbf{0.21}  \\ 
    \hline
    \end{tabular}}
    \caption{Performance comparison of image synthesis from novel views between the proposed method and additional baselines on the nuScenes dataset.}
    \label{tab:table6}
    \end{minipage}
\end{table}
\begin{table*}[t]
\centering
\resizebox{0.999\textwidth}{!}{
\begin{tabular}{C{5.6cm} C{1.5cm} C{1.5cm} C{1.5cm} C{1.5cm} C{1.5cm} C{1.5cm}}
\hline
\textbf{Method}   & PSNR↑ & SSIM↑ & LPIPS↓ & AbsRel↓ & RMSE↓ & RMSElog↓    \\  
\hline
TCLC-GS w/o 3D mesh        & 26.36 & 0.82 & 0.28  & 0.42 & 10.92 & 1.06  \\
TCLC-GS w/o colorized 3D mesh        & 27.61  & 0.85  & 0.22 & 0.03  & 1.55 & 0.05\\
TCLC-GS w/o octree implicit feature        & 27.81  & 0.85  & 0.23  & 0.04 & 1.63 & 0.05 \\
TCLC-GS w/o dense depth supervision        & 27.96  & 0.86  & 0.22 & 0.37 & 9.80 & 0.35 \\ 
TCLC-GS full        & \textbf{28.11} & \textbf{0.86} & \textbf{0.22} & \textbf{0.03} & \textbf{1.54} & \textbf{0.05} \\ 
\hline
\end{tabular}}
\caption{Ablation study of the proposed method on the Waymo dataset.}
\label{tab:table7}
\end{table*}
\begin{table}[t]
    \centering
    \begin{minipage}{0.48\textwidth}
        \centering
        \resizebox{1.0\textwidth}{!}{
        \begin{tabular}{C{3cm} C{3cm} }
        \hline
        \textbf{Method}   & FPS  \\  
        \hline
        SUDS ~\cite{turki2023suds}                        & $\sim$ 0.01  \\
        StreetSurf~\cite{guo2023streetsurf}                    & $\sim$ 0.10  \\
        S-NeRF~\cite{li2024scenarionet}                       & $\sim$ 0.02  \\
        3D-GS~\cite{kerbl20233d}     & $\sim$ 200  \\
        \hline
        TCLC-GS                      & $\sim$ 90   \\ 
        \hline
        \end{tabular}}
        \caption{The comparison running-time analysis on the Waymo dataset.}
        \label{tab:table7}
    \end{minipage}
    \hfill
    \begin{minipage}{0.48\textwidth}
        \resizebox{1.0\textwidth}{!}{
        \begin{tabular}{C{3cm} C{3cm}}
        \hline
        \textbf{Method}   & FPS    \\  
        \hline
        Instant-NGP~\cite{muller2022instant}         & $\sim$ 0.23  \\
        Mip-NeRF360~\cite{barron2022mip}             & $\sim$ 0.08  \\
        Urban-NeRF~\cite{rematas2022urban}           & $\sim$ 0.03  \\
        SUDS~\cite{turki2023suds}                    & $\sim$ 0.02  \\
        S-NeRF~\cite{li2024scenarionet}              & $\sim$ 0.04  \\
        3D-GS~\cite{kerbl20233d}     & $\sim$ 350     \\
        \hline
        TCLC-GS                      & $\sim$ 120     \\ 
        \hline
        \end{tabular}}
        \caption{The comparison running-time analysis on the nuScenes dataset.}
        \label{tab:table8}
    \end{minipage}
\end{table}

\textbf{Ablation Study:}
To demonstrate the individual effectiveness of each component in our method, we conducted ablation studies using the Waymo dataset. 
Since our contributions mainly include colorized 3D mesh, octree implicit representation, and dense depth supervision, our ablation study analyses the impact of our designs from these aspects.
We train five different variations: 
1) TCLC-GS without 3D mesh, initializing 3D Gaussian using original 3D LiDAR points;
2) TCLC-GS without colorized 3D mesh, initializing 3D Gaussian using 3D mesh without color information;
3) TCLC-GS without octree implicit representation, initializing 3D Gaussian using colorized 3D mesh without octree representation;
4) TCLC-GS without dense depth supervision, training TCLC-GS only using RGB supervision;
5) TCLC-GS full method.
6) TCLC-GS + sky refine
The average values of the evaluation metrics across six testing scenes from the Waymo dataset are given in Table~\ref{tab:table7}.
These ablation results indirectly validate the effectiveness of colorized 3D mesh, octree implicit representation, and dense depth supervision, underscoring their contributions to the overall performance.

\textbf{Running-time Analysis:}
We present a running time performance comparison between our TCLC-GS and various baselines on the Waymo and nuScenes datasets, detailed in Table~\ref{tab:table7} and Table~\ref{tab:table8}, respectively.
Utilizing a single NVIDIA GeForce RTX 3090 Ti, TCLC-GS attains real-time rendering speeds with high-resolution images, achieving around 90 FPS with an image resolution of 1920$\times$1280 on the Waymo dataset, and around 120 FPS with an image resolution of 1600$\times$900 on the nuScenes dataset.
Compared to NeRF-based methods such as SUDS~\cite{turki2023suds} and S-NeRF~\cite{li2024scenarionet}, TCLC-GS significantly outperforms the speed of these methods, achieving real-time running performance. 
When compared to 3D-GS~\cite{kerbl20233d}, TCLC-GS offers higher accuracy in both RGB and depth rendering, and still maintains the performance efficiency.
\section{Conclusion}
\label{sec:conclusion}
In this paper, we proposed a novel Tightly Coupled LiDAR-Camera Gaussian Splatting~(TCLC-GS) that synergizes the strengths of LiDAR and surrounding cameras for fast modeling and real-time rendering in urban driving scenarios. 
The key idea of TCLC-GS is a hybrid 3D representation combining explicit (colorized 3D mesh) and implicit (hierarchical octree feature) information derived from the LiDAR-camera data, which enriches both geometry and appearance properties of 3D Gaussians. 
The optimization of Gaussian Splatting is further enhanced by incorporating rendered dense depth data within the 3D mesh.
The experimental evaluations demonstrate that our model surpasses SOTA performance while maintaining the real-time efficiency of Gaussian Splatting on Waymo Open and nuScenes datasets.
\newpage
In this Appendix, we firstly provide more details of our TCLC-GS implementation, including network architecture, and hyper-parameters of octree feature grid and 3D Gaussians in Appendix~\ref{sec:implementation}. 
Secondly, more comparison visualizations with the baseline are presented in Appendix~\ref{sec:visualization}.
Lastly, three video demos are attached in Appendix~\ref{sec:demo}.

\section{Implementation Details}
\label{sec:implementation}

\begin{figure*}[h]
    \centering
    \includegraphics[width=1.0\textwidth]{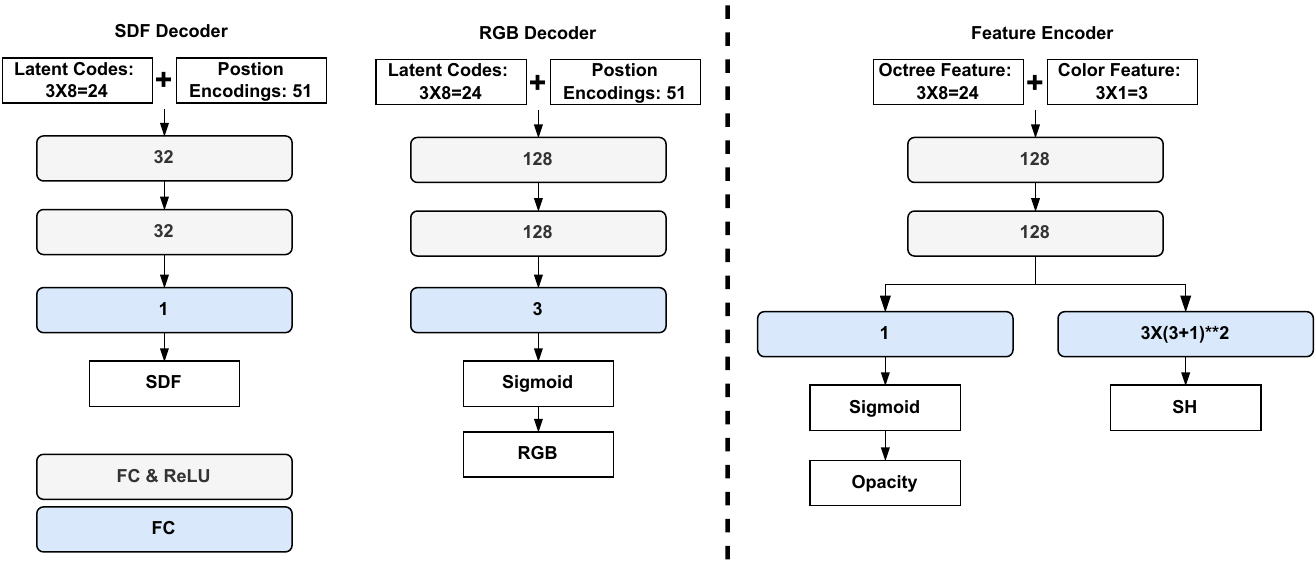}
    \caption{Left: network architecture of SDF and RGB decoders in octree training.
    Right: network architecture of feature encoder for Gaussian splatting.}
  \label{fig:network_details}
\end{figure*}

\textbf{Network Architecture:}
Figure~\ref{fig:network_details} illustrates the network architecture of the SDF and RGB decoders in octree training, and feature encoder for Gaussian splatting. 
For building the hierarchical octree feature and colorized 3D mesh,
the SDF and RGB decoders employ a shallow multilayer perceptron (MLP) consisting of three fully connected layers. 
Each of these layers is followed by a ReLU activation, except for the final prediction layer. 
A Sigmoid layer is inserted after the final layer in the RGB decoder.
In terms of the optimization of 3D Gaussians,
the feature encoder adopts a shared two layers MLP following a dual-brunch  fully connected layer.
Each of MLP is followed by a ReLU activation, except for the final prediction layer.
A Sigmoid layer is inserted after the final layer in the opacity prediction.

\textbf{Octree Feature Grid Hyper-parameters:}
We construct the octree with a height of 12, storing the latent features in the last three layers.
The leaf voxel resolution in our configuration is set to 0.2m, and the latent code assigned to each corner of the octree node is set to 8 dimension.

\textbf{3D Gaussian Hyper-parameters:}
We set the position learning rate to vary from 1.6e-5 to 1.6e-6,
set opacity learning rate as 0.05,
scale learning rate as 0.01,
feature learning rate a 2.5e-3,
and rotation learning rate as  0.001.
We configure the interval of densification and opacity reset as 500 and 3000 respectively. 
We set the densify grad threshold as 2e-4.
Additionally, the Spherical Harmonics (SH) degree for each 3D Gaussian is established at 3.

\section{More Visualization Results}
\label{sec:visualization}

\begin{figure*}[!t]
    \centering
    \includegraphics[width=1.0\textwidth]{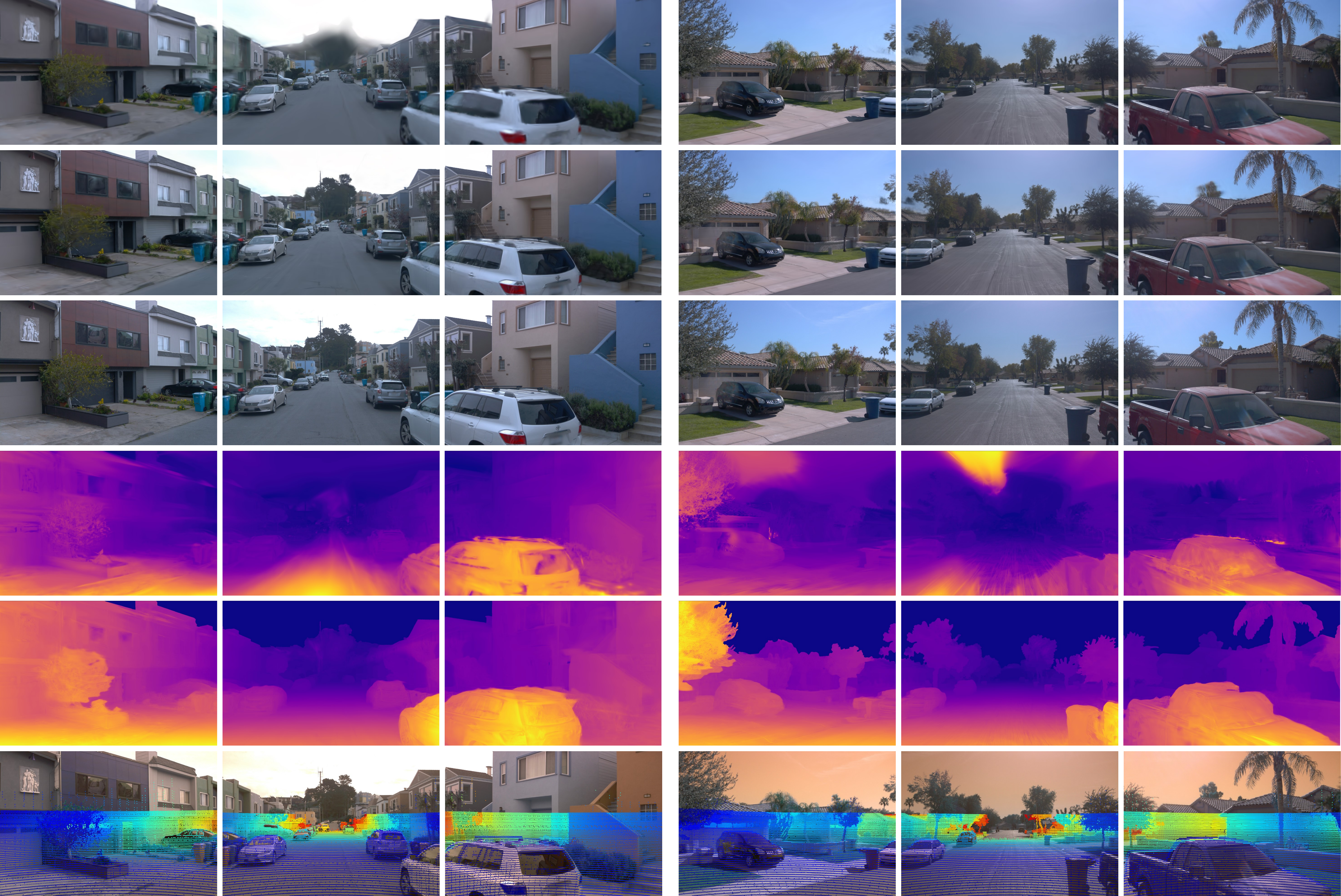}
    \caption{Visual comparison of image and depth synthesis from novel front-left, front, and front-right surrounding views on the Waymo dataset. 
    Row 1: 3D-GS images;
    Row 2: TCLC-GS images;
    Row 3: GT images;
    Row 4: 3D-GS depths;
    Row 5: TCLC-GS depths;
    Row 6: GT depth of LiDAR points projected on images.}
  \label{fig:waymo_vis2}
\end{figure*}

\begin{figure*}[!t]
    \centering
    \includegraphics[width=1.0\textwidth]{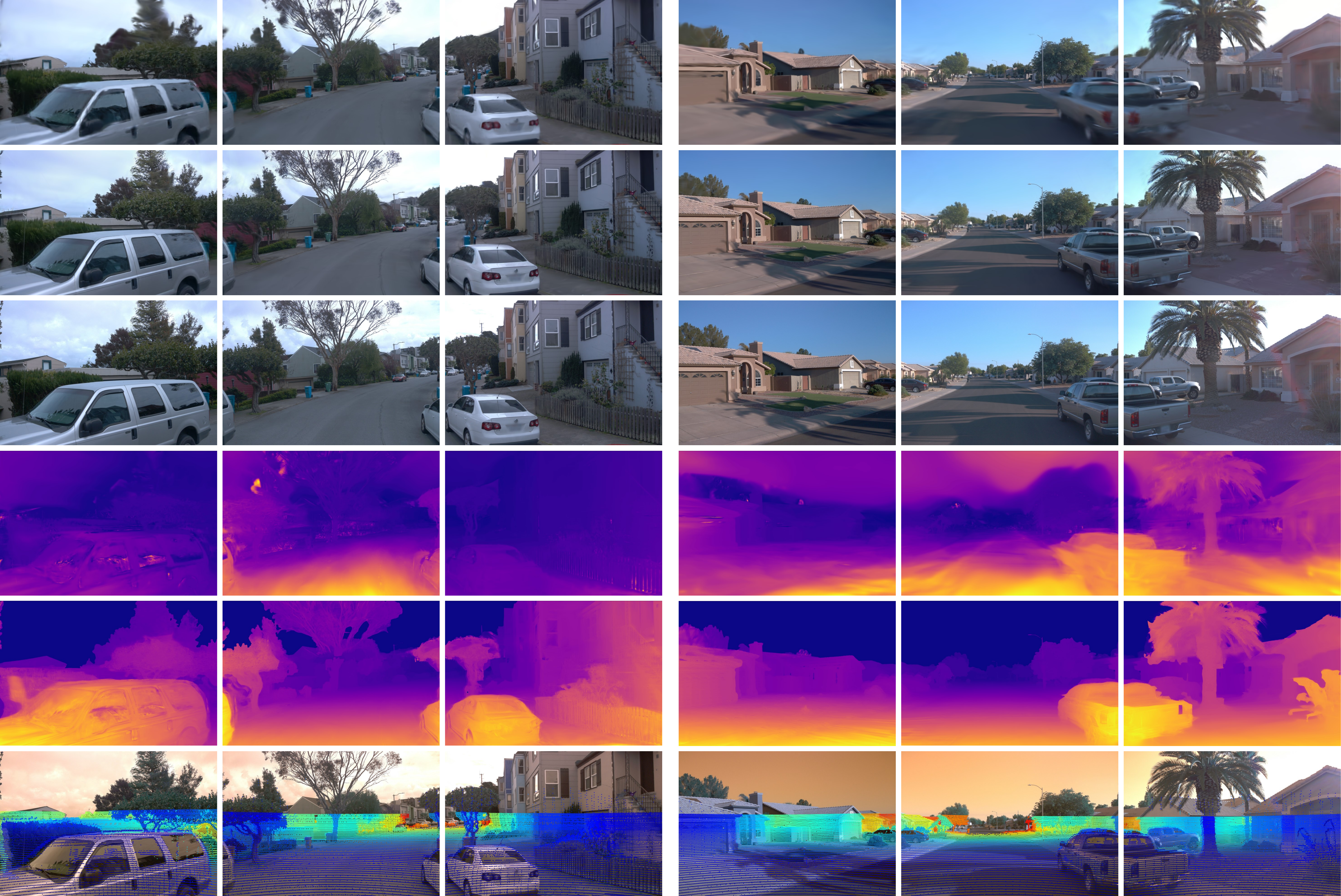}
    \caption{Visual comparison of image and depth synthesis from novel front-left, front, and front-right surrounding views on the Waymo dataset. 
    Row 1: 3D-GS images;
    Row 2: TCLC-GS images;
    Row 3: GT images;
    Row 4: 3D-GS depths;
    Row 5: TCLC-GS depths;
    Row 6: GT depth of LiDAR points projected on images.}
  \label{fig:waymo_vis3}
\end{figure*}

\textbf{Visualization Results on the Waymo Open dataset:}
We present visual comparison results of image and depth synthesis on the Waymo dataset in Fig.~\ref{fig:waymo_vis2} and Fig.~\ref{fig:waymo_vis3}. 
It can be seen that TCLC-GS produces clearer images compared to 3D-GS~\cite{kerbl20233d}. Additionally, the depth images generated by TCLC-GS are denser and sharper than those produced by 3D-GS.

\begin{figure*}[!t]
    \centering
    \includegraphics[width=1.0\textwidth]{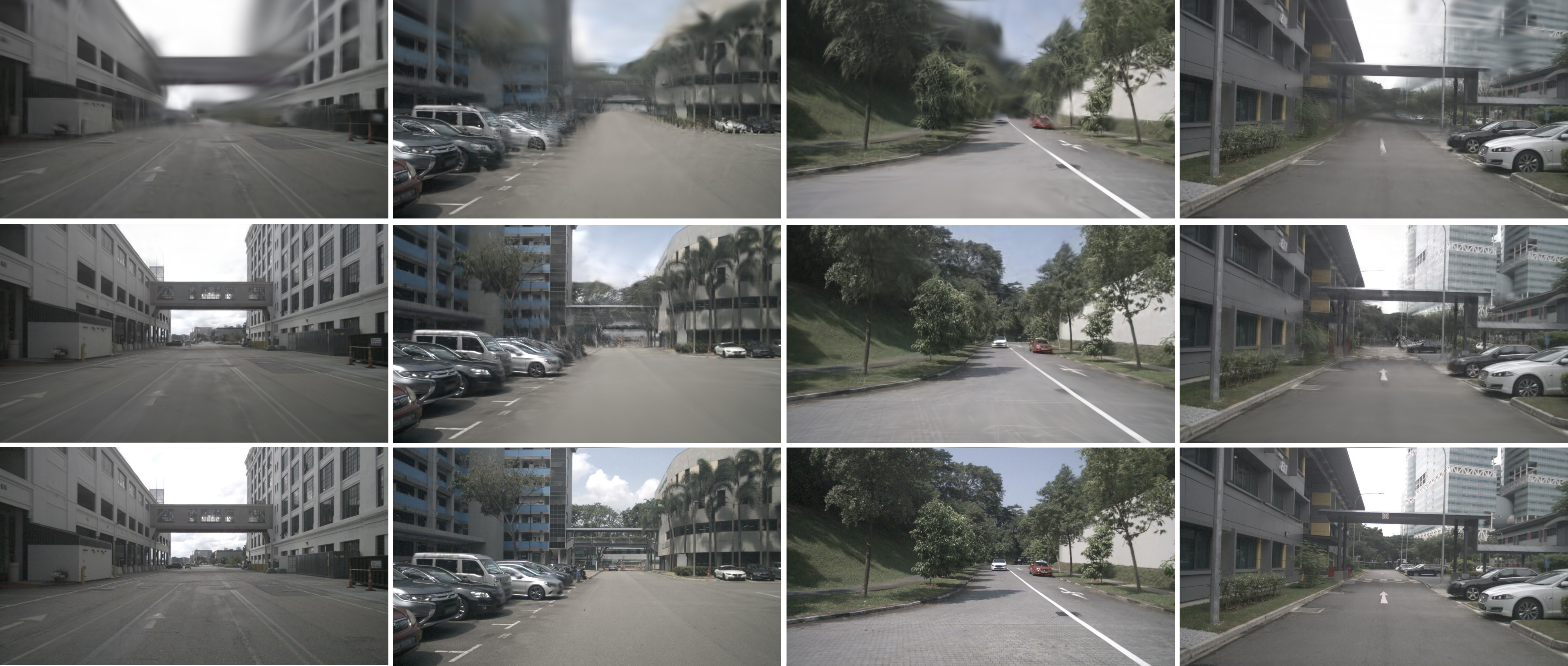}
    \caption{Visual comparison of image synthesis from novel views on the nuScenes dataset. 
    Row 1: 3D-GS;
    Row 2: TCLC-GS;
    Row 3: GT.}
  \label{fig:nuScenes_image1}
\end{figure*}

\begin{figure*}[!t]
    \centering
    \includegraphics[width=1.0\textwidth]{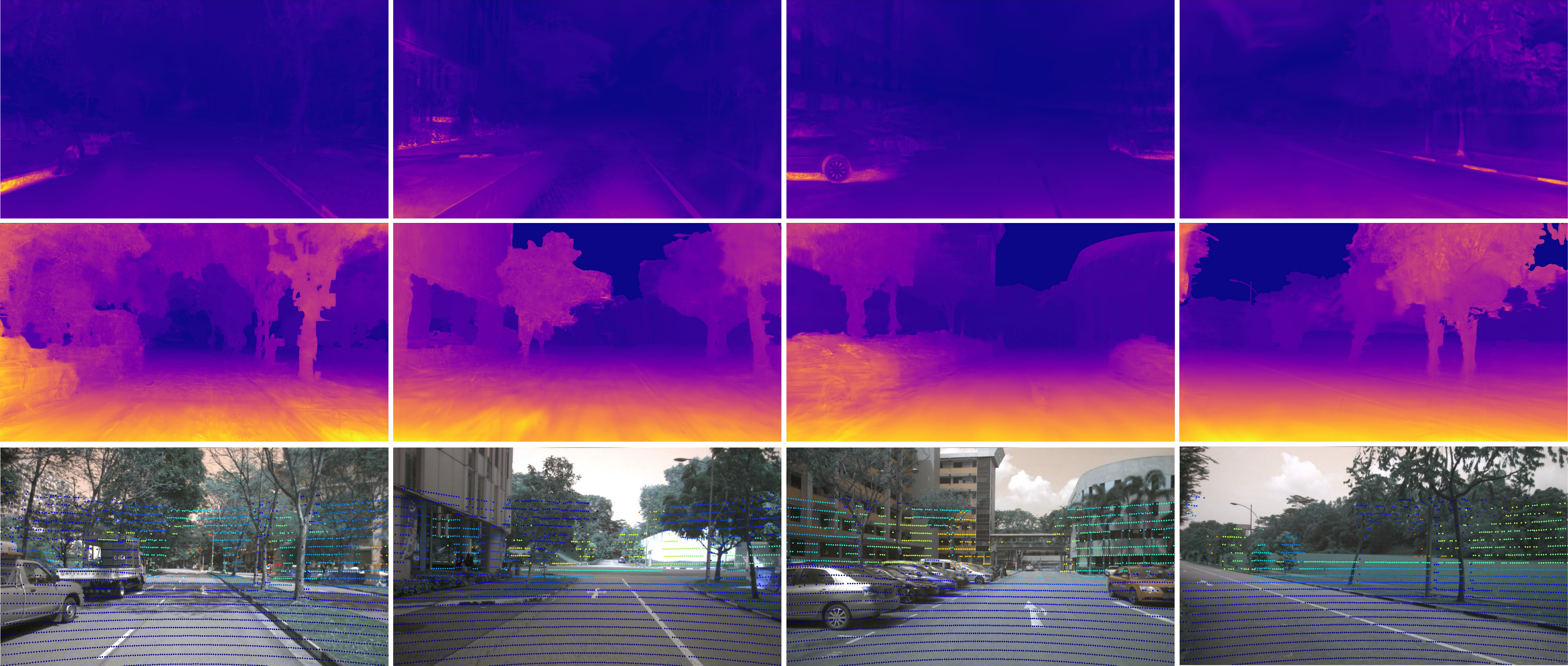}
    \caption{Visual comparison of depth synthesis from novel views on the nuScenes dataset. 
    Row 1: 3D-GS;
    Row 2: TCLC-GS;
    Row 3: GT of LiDAR points projected on images.}
  \label{fig:nuScenes_depth1}
\end{figure*}

\textbf{Visualization Results on the nuScenes dataset:}
We present visual comparison results of image and depth synthesis on the nuScenes dataset in Fig.~\ref{fig:nuScenes_image1} and Fig.~\ref{fig:nuScenes_depth1} separately. 
It's observable that the images generated by TCLC-GS exhibit fewer blurring areas compared to those from 3D-GS~\cite{kerbl20233d}. 
The depth images rendered by TCLC-GS are denser compared to those produced by 3D-GS.

\section{Video Demo}
\label{sec:demo}
We provide Video Demo 1, Demo 2 and Demo 3 in the website~\footnote{\url{https://github.com/BoschRHI3NA/Video-Demo-ECCV24/tree/main}}. 
Demos 1 and 2 illustrate the rendered images and depths by our TCLC-GS, alongside the ground truth (GT) data, in two different urban scenes.
In these videos, the first row displays the images rendered by TCLC-GS, the second row shows the GT images, the third row presents the dense depth rendered by TCLC-GS, and the fourth row illustrates the GT sparse depth obtained by projecting LiDAR data onto the surrounding images.
Demo 3 showcases a more challenging rendering scenario.
This video presents image and depth rendering from a new generated ego-car trajectory which is 0.5 meter higher or 0.5 meter lower than the original ego-car trajectory. 
The first and third rows display the rendered RGB and depth images from a trajectory 0.5m lower than the original, while the second and fourth rows show the rendered RGB and depth images from a trajectory 0.5m higher than the original.
This demonstration highlights our method's potential to address the domain gap in data reuse caused by different hardware car settings, such as collecting data from a sedan but deploying it on an SUV.
%
%

\par\vfill\par


%
%
\bibliographystyle{splncs04}
\bibliography{main}

\begin{thebibliography}{10}
\providecommand{\url}[1]{\texttt{#1}}
\providecommand{\urlprefix}{URL }
\providecommand{\doi}[1]{https://doi.org/#1}

\bibitem{barron2021mip}
Barron, J.T., Mildenhall, B., Tancik, M., Hedman, P., Martin-Brualla, R., Srinivasan, P.P.: Mip-nerf: A multiscale representation for anti-aliasing neural radiance fields. In: Proceedings of the IEEE/CVF International Conference on Computer Vision. pp. 5855--5864 (2021)

\bibitem{barron2022mip}
Barron, J.T., Mildenhall, B., Verbin, D., Srinivasan, P.P., Hedman, P.: Mip-nerf 360: Unbounded anti-aliased neural radiance fields. In: Proceedings of the IEEE/CVF Conference on Computer Vision and Pattern Recognition. pp. 5470--5479 (2022)

\bibitem{bhat2023zoedepth}
Bhat, S.F., Birkl, R., Wofk, D., Wonka, P., M{\"u}ller, M.: Zoedepth: Zero-shot transfer by combining relative and metric depth. arXiv preprint arXiv:2302.12288  (2023)

\bibitem{caesar2020nuscenes}
Caesar, H., Bankiti, V., Lang, A.H., Vora, S., Liong, V.E., Xu, Q., Krishnan, A., Pan, Y., Baldan, G., Beijbom, O.: nuscenes: A multimodal dataset for autonomous driving. In: Proceedings of the IEEE/CVF conference on computer vision and pattern recognition. pp. 11621--11631 (2020)

\bibitem{chen2023periodic}
Chen, Y., Gu, C., Jiang, J., Zhu, X., Zhang, L.: Periodic vibration gaussian: Dynamic urban scene reconstruction and real-time rendering. arXiv preprint arXiv:2311.18561  (2023)

\bibitem{glassner1989introduction}
Glassner, A.S.: An introduction to ray tracing. Morgan Kaufmann (1989)

\bibitem{monodepth17}
Godard, C., {Mac Aodha}, O., Brostow, G.J.: Unsupervised monocular depth estimation with left-right consistency. In: CVPR (2017)

\bibitem{icml2020_2086}
Gropp, A., Yariv, L., Haim, N., Atzmon, M., Lipman, Y.: Implicit geometric regularization for learning shapes. In: Proceedings of Machine Learning and Systems 2020, pp. 3569--3579 (2020)

\bibitem{guedon2023sugar}
Gu{\'e}don, A., Lepetit, V.: Sugar: Surface-aligned gaussian splatting for efficient 3d mesh reconstruction and high-quality mesh rendering. In: Proceedings of the IEEE/CVF Conference on Computer Vision and Pattern Recognition (2024)

\bibitem{guo2023streetsurf}
Guo, J., Deng, N., Li, X., Bai, Y., Shi, B., Wang, C., Ding, C., Wang, D., Li, Y.: Streetsurf: Extending multi-view implicit surface reconstruction to street views. arXiv preprint arXiv:2306.04988  (2023)

\bibitem{kerbl20233d}
Kerbl, B., Kopanas, G., Leimk{\"u}hler, T., Drettakis, G.: 3d gaussian splatting for real-time radiance field rendering. ACM Transactions on Graphics  \textbf{42}(4) (2023)

\bibitem{li2024scenarionet}
Li, Q., Peng, Z.M., Feng, L., Liu, Z., Duan, C., Mo, W., Zhou, B.: Scenarionet: Open-source platform for large-scale traffic scenario simulation and modeling. Advances in neural information processing systems  \textbf{36} (2024)

\bibitem{liu2023real}
Liu, J.Y., Chen, Y., Yang, Z., Wang, J., Manivasagam, S., Urtasun, R.: Real-time neural rasterization for large scenes. In: Proceedings of the IEEE/CVF International Conference on Computer Vision. pp. 8416--8427 (2023)

\bibitem{lorensen1987marching}
Lorensen, W.E., Cline, H.E.: Marching cubes: A high resolution 3d surface construction algorithm. ACM siggraph computer graphics  \textbf{21}(4),  163--169 (1987)

\bibitem{lu2023urban}
Lu, F., Xu, Y., Chen, G., Li, H., Lin, K.Y., Jiang, C.: Urban radiance field representation with deformable neural mesh primitives. In: Proceedings of the IEEE/CVF International Conference on Computer Vision. pp. 465--476 (2023)

\bibitem{martin2021nerf}
Martin-Brualla, R., Radwan, N., Sajjadi, M.S., Barron, J.T., Dosovitskiy, A., Duckworth, D.: Nerf in the wild: Neural radiance fields for unconstrained photo collections. In: Proceedings of the IEEE/CVF Conference on Computer Vision and Pattern Recognition. pp. 7210--7219 (2021)

\bibitem{mildenhall2021nerf}
Mildenhall, B., Srinivasan, P.P., Tancik, M., Barron, J.T., Ramamoorthi, R., Ng, R.: Nerf: Representing scenes as neural radiance fields for view synthesis. Communications of the ACM  \textbf{65}(1),  99--106 (2021)

\bibitem{muller2022instant}
M{\"u}ller, T., Evans, A., Schied, C., Keller, A.: Instant neural graphics primitives with a multiresolution hash encoding. ACM Transactions on Graphics (ToG)  \textbf{41}(4),  1--15 (2022)

\bibitem{Ortiz:etal:iSDF2022}
Ortiz, J., Clegg, A., Dong, J., Sucar, E., Novotny, D., Zollhoefer, M., Mukadam, M.: isdf: Real-time neural signed distance fields for robot perception. In: Robotics: Science and Systems (2022)

\bibitem{ost2022neural}
Ost, J., Laradji, I., Newell, A., Bahat, Y., Heide, F.: Neural point light fields. In: Proceedings of the IEEE/CVF Conference on Computer Vision and Pattern Recognition. pp. 18419--18429 (2022)

\bibitem{rematas2022urban}
Rematas, K., Liu, A., Srinivasan, P.P., Barron, J.T., Tagliasacchi, A., Funkhouser, T., Ferrari, V.: Urban radiance fields. In: Proceedings of the IEEE/CVF Conference on Computer Vision and Pattern Recognition. pp. 12932--12942 (2022)

\bibitem{sun2020scalability}
Sun, P., Kretzschmar, H., Dotiwalla, X., Chouard, A., Patnaik, V., Tsui, P., Guo, J., Zhou, Y., Chai, Y., Caine, B., et~al.: Scalability in perception for autonomous driving: Waymo open dataset. In: Proceedings of the IEEE/CVF conference on computer vision and pattern recognition. pp. 2446--2454 (2020)

\bibitem{takikawa2021nglod}
Takikawa, T., Litalien, J., Yin, K., Kreis, K., Loop, C., Nowrouzezahrai, D., Jacobson, A., McGuire, M., Fidler, S.: Neural geometric level of detail: Real-time rendering with implicit {3D} shapes  (2021)

\bibitem{tancik2022block}
Tancik, M., Casser, V., Yan, X., Pradhan, S., Mildenhall, B., Srinivasan, P.P., Barron, J.T., Kretzschmar, H.: Block-nerf: Scalable large scene neural view synthesis. In: Proceedings of the IEEE/CVF Conference on Computer Vision and Pattern Recognition. pp. 8248--8258 (2022)

\bibitem{turki2023suds}
Turki, H., Zhang, J.Y., Ferroni, F., Ramanan, D.: Suds: Scalable urban dynamic scenes. In: Proceedings of the IEEE/CVF Conference on Computer Vision and Pattern Recognition. pp. 12375--12385 (2023)

\bibitem{xie2021segformer}
Xie, E., Wang, W., Yu, Z., Anandkumar, A., Alvarez, J.M., Luo, P.: Segformer: Simple and efficient design for semantic segmentation with transformers. In: Neural Information Processing Systems (NeurIPS) (2021)

\bibitem{xu2022point}
Xu, Q., Xu, Z., Philip, J., Bi, S., Shu, Z., Sunkavalli, K., Neumann, U.: Point-nerf: Point-based neural radiance fields. In: Proceedings of the IEEE/CVF Conference on Computer Vision and Pattern Recognition. pp. 5438--5448 (2022)

\bibitem{yan2024street}
Yan, Y., Lin, H., Zhou, C., Wang, W., Sun, H., Zhan, K., Lang, X., Zhou, X., Peng, S.: Street gaussians for modeling dynamic urban scenes. arXiv preprint arXiv:2401.01339  (2024)

\bibitem{yang2023emernerf}
Yang, J., Ivanovic, B., Litany, O., Weng, X., Kim, S.W., Li, B., Che, T., Xu, D., Fidler, S., Pavone, M., et~al.: Emernerf: Emergent spatial-temporal scene decomposition via self-supervision. arXiv preprint arXiv:2311.02077  (2023)

\bibitem{yang2024depth}
Yang, L., Kang, B., Huang, Z., Xu, X., Feng, J., Zhao, H.: Depth anything: Unleashing the power of large-scale unlabeled data. arXiv preprint arXiv:2401.10891  (2024)

\bibitem{zhou2023drivinggaussian}
Zhou, X., Lin, Z., Shan, X., Wang, Y., Sun, D., Yang, M.H.: Drivinggaussian: Composite gaussian splatting for surrounding dynamic autonomous driving scenes. arXiv preprint arXiv:2312.07920  (2023)

\end{thebibliography}
\end{document}